\pdfoutput=1

\documentclass[11pt]{article}

\usepackage{ACL2023}

\usepackage{times}
\usepackage{latexsym}

\usepackage[T1]{fontenc}

\usepackage[utf8]{inputenc}

\usepackage{microtype}

\usepackage{inconsolata}

\usepackage{anyfontsize}
\usepackage{amsmath}
\usepackage{array}
\usepackage{booktabs}
\usepackage{CJKutf8}
\usepackage[inline,shortlabels]{enumitem}
\usepackage{graphicx}
\usepackage{multirow}
\usepackage{pgfplots}
\usepackage{pgfplotstable}
\usepackage{soul}
\usepackage{subcaption}
\usepackage{trimclip}
\usepackage{xspace}

\pgfplotsset{compat=1.16}
\usepgfplotslibrary{statistics}
\usepgfplotslibrary{groupplots}

\newif\ifblind
\blindfalse

\ifblind
\newcommand{\palm}{BigLM\xspace}
\newcommand{\gtrans}{Google Translate\xspace}
\newcommand{\gtransshort}{Google Trans.\xspace}
\newcommand{\blindcite}[1]{(Blind Cite, 20XX)}
\newcommand{\blindcitet}[1]{Blind Cite (20XX)}
\newcommand{\blindcitep}[1]{(Blind Cite, 20XX)}
\else
\newcommand{\palm}{PaLM\xspace}
\newcommand{\gtrans}{Google Translate\xspace}
\newcommand{\gtransshort}{Google Trans.\xspace}
\newcommand{\blindcite}[1]{\cite{#1}}
\newcommand{\blindcitet}[1]{\citet{#1}}
\newcommand{\blindcitep}[1]{\citep{#1}}
\fi

\newcommand{\kNN}{$k\text{NN}$\xspace}
\newcommand{\wmtdev}{WMT-dev\xspace}
\newcommand{\wmtfull}{WMT-full\xspace}
\newcommand{\highend}{high-end\xspace}
\newcommand{\langdir}[2]{#1 $\rightarrow$ #2\xspace}
\newcommand{\langdirtight}[2]{#1$\rightarrow$#2\xspace}
\newcommand{\langbidir}[2]{#1 $\leftrightarrow$ #2\xspace}
\newcommand{\langbidirtight}[2]{#1$\leftrightarrow$#2\xspace}
\newcommand{\roberta}{Ro\textsc{bert}a\xspace}
\newcommand{\bleurt}{\textsc{bleurt}\xspace}
\newcommand{\bleu}{\textsc{bleu}\xspace}
\newcommand{\comet}{\textsc{comet}\xspace}
\newcommand{\mqm}{\textsc{mqm}\xspace}
\newcommand{\sacrebleu}{\textsc{sacrebleu}\xspace}

\newcommand{\zh}[1]{\begin{CJK*}{UTF8}{gbsn}#1\end{CJK*}}
\newcolumntype{L}[1]{>{\raggedright\let\newline\\\arraybackslash\hspace{0pt}}p{#1}}
\newcolumntype{R}[1]{>{\raggedleft\let\newline\\\arraybackslash\hspace{0pt}}m{#1}}

\definecolor{shadecolor}{RGB}{200,200,200}
\definecolor{minorcolor}{RGB}{251,236,93}

\newcommand{\major}[1]{{\setlength{\fboxsep}{0pt}\colorbox{pink}{#1}}}
\newcommand{\minor}[1]{{\setlength{\fboxsep}{0pt}\colorbox{minorcolor}{#1}}}

\newcolumntype{H}{>{\setbox0=\hbox\bgroup}c<{\egroup}@{}}

\colorlet{graphGreen}{green!40!teal!90}
\colorlet{graphBlue}{blue!90!teal!20}

\makeatletter
\g@addto@macro\bfseries{\boldmath}
\makeatother

\newcommand{\todo}[2][all]{}  

\title{Prompting \palm for Translation: Assessing Strategies and Performance}

\author{David Vilar, Markus Freitag, Colin Cherry, Jiaming Luo, Viresh Ratnakar, George Foster\\
  Google \\
\texttt{\{vilar, freitag, colincherry, jmluo, vratnakar, fosterg\}@google.com} \\}

\begin{document}
\maketitle
\begin{abstract}
Large language models (LLMs) that have been trained on multilingual but not parallel text exhibit a remarkable ability to translate between languages. We probe this ability in an in-depth study of the pathways language model (\palm), which has demonstrated the strongest machine translation (MT) performance among similarly-trained LLMs to date. We investigate various strategies for choosing translation examples for few-shot prompting, concluding that example quality is the most important factor. Using optimized prompts, we revisit previous assessments of {\palm}’s MT capabilities with more recent test sets, modern MT metrics, and human evaluation, and find that its performance, while impressive, still lags that of state-of-the-art supervised systems. We conclude by providing an analysis of {\palm}’s MT output which reveals some interesting properties and prospects for future work.
\end{abstract}

\section{Introduction}

Large language models (LLMs) trained to predict the next token from a lengthy context have demonstrated impressive machine translation capabilities, despite being trained on corpora that are overwhelmingly English, with no intentionally-included parallel text. 
In this paper, we carry out an in-depth investigation into the translation capabilities of LLMs, 
testing different prompting strategies and carefully assessing the resulting performance.
We study the recently-introduced \palm model \blindcite{chowdhery2022palm}, a 540B-parameter decoder-only language model trained on a heavily English-centric, multilingual corpus.
It has achieved the strongest MT results among LLMs trained on non-parallel multilingual corpora. 

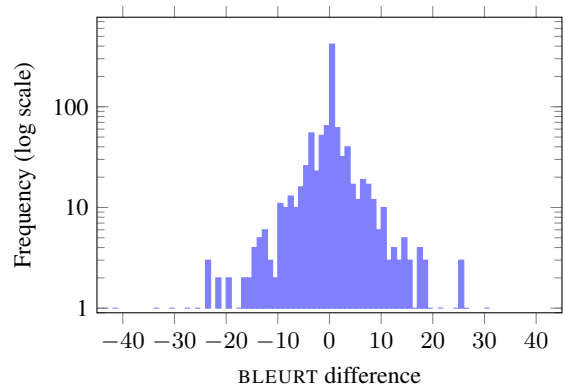
\begin{figure}[t]
    \centering

    \begin{tikzpicture}
        \small
        \begin{axis}[
            ybar,
            bar width=1,
            ymode=log,
            yticklabels={0, 1, 10, 100},
            width=\columnwidth,
            height=5.5cm,
            xlabel={\bleurt difference},
            ylabel={Frequency (log scale)},
            xtick={-40, -30,...,40},
            xmin=-45, xmax=45,
            ymin=0.9,
            clip=false,
        ]
        \addplot [fill=blue!50, color=blue!50] table {bleurtScores/bleurt.random1-random3.5shot.histWidth1};
        \end{axis}
    \end{tikzpicture}
    
    \caption{
        Histogram of the sentence-level \bleurt difference between two different 5-shot \palm runs using the random prompt selection method from the original paper on a corpus of 1000 sentences.
        Each bar corresponds to a difference range of 1 \bleurt point. A majority of sentences (516) show a difference of more than 1 \bleurt point, demonstrating that the choice of prompt can strongly affect translation quality. 
   }
    
    \label{fig:random1_vs_random2}
\end{figure}

To ensure a fair assessment of {\palm}'s MT capability, we begin with an exploration of example selection methods for use with fixed prompt templates.
We vary both the pool from which examples are chosen and the method for choosing them, comparing standard random selection to $k$-nearest-neighbour~(\kNN) selection that customizes prompts for specific inputs. Figure~\ref{fig:random1_vs_random2} highlights the importance of example selection by showing that two randomly-selected sets of examples can result in significantly different distributions of sentence-level \bleurt scores.

Although \citet{chowdhery2022palm} report interesting results on low-resource and non-English language pairs, their most striking findings concern high-resource pairs. Accordingly, we limit our investigation to French, German, and Chinese translation to and from English.
We evaluate sentence-level translation quality using recommended practices for high-quality MT, specifically:  \begin{enumerate*}[(i)] \item we use recent WMT test sets to guard against train/test data leakage, and to facilitate comparison with state-of-the-art~(SOTA) MT systems; \item we use a SOTA automatic metric (\bleurt) instead of \bleu which has been demonstrated to be suboptimal for high-quality translations ~\cite{kocmi-etal-2021-ship, freitag-etal-2021-results}; and \item we conduct an expert-based human evaluation with detailed categories to characterize the error patterns of the automatically generated translations. \end{enumerate*}

Our contributions are as follows:
\begin{itemize}
    \item We carry out the first systematic study of LLM prompting for MT, exploring both the example candidate pool and the selection strategy. We find that the \emph{quality} of examples matters more than the domain from which they are drawn or their lexico-semantic proximity to the current input.
    \item We evaluate the translation capability of LLMs with the procedure currently recommended by the MT community.
    We find that, although impressive, the sentence-level translation capacity of LLMs still lags behind SOTA MT.
\end{itemize}

\section{Related Work}

Inspired by the findings of \citet{radford2019language,brown2020language}, prompting strategies for LLMs have become a topic of intense interest, generating work across a broad spectrum of methods and applications \cite{liu2021pre}. A basic distinction can be made between \emph{hard} (explicit text) prompting such as we use, and \emph{soft} prompting that seeks to learn embeddings \cite{lester-etal-2021-power}, activations \cite{li-liang-2021-prefix,hambardzumyan-etal-2021-warp}, or attention weights \cite{liu2022few} that condition the model to perform a desired task. The latter approach is more expressive and more efficient at inference time, but performance can be sensitive to initialization \cite{hou2022metaprompting}, and some techniques require modifications to the model.

Hard prompts have the advantage of being easy to interpret and modify.
Work in this area includes tools to facilitate development of handcrafted prompts \cite{strobelt2022interactive,bach2022promptsource}; algorithms to find optimal prompts through gradient-guided search \cite{shin2020autoprompt} or exhaustive search through labels \cite{schick2021exploiting} or both labels and templates \cite{gao2021making}; as well as studies on the effect of example order \cite{kumar2021reordering,lu-etal-2022-fantastically}.
Hard prompts have also been used to analyze model capabilities \cite{garg2022can,li2022probing}, the role of data \cite{singh2022explaining}, and the nature of prompting itself \cite{min2022rethinking,wei2022chain}.


With few exceptions, e.g. \cite{li-etal-2022-learning-transfer,liu-etal-2022-makes,valvoda2022prompting}, early approaches to hard prompting tended to condition on the task rather than the specific input.
Our \kNN approach for conditioning on the input was pioneered by \citet{liu-etal-2022-makes}, who used RoBERTa embeddings to identify relevant GPT-3 prompts for sentiment, table-to-text, and QA tasks.
They found that \kNN works better than a random-selection baseline, and that the advantage grows as the size of the (domain-controlled) example pool increases.

Work on prompting LLMs for MT began with the GPT-3 and \palm papers \cite{brown2020language,chowdhery2022palm}, which adopted similar approaches, comparing 0, 1, and $n$-shot\footnote{Where $n$ is 64 for GPT-3 and 5 for \palm.} random selection of independent sentence pairs from WMT training corpora, and testing on older French, German, and Romanian WMT test sets traditionally used in ML, augmented in \palm with \langdirtight{French}{German} and Kazakh.
For both models, performance increased with number of shots, and $n$-shot \bleu scores were found to be competitive with previous unsupervised SOTA, and in some settings---particularly into English---supervised SOTA as well.

In other early MT work, \citet{reynolds2021prompt} experimented with prompt templates for GPT-3, and found that 0-shot prompts with carefully-chosen templates can outperform $n$-shot prompts with sub-optimal templates.
\citet{garcia2022using} explored using prompts with mT5 \cite{xue-etal-2021-mt5} to control output attributes such as formality, and also examine the effect of using prompt-like natural-language tags during fine-tuning.
\citet{patel2022bidirectional} proposed autoregressive prompting: concatenating only the first predicted word to a prompt and output prefix at each step.

\subsection*{Après nous, le déluge}

Since our paper appeared on arXiv in November 2022, there has been a flood of work on using LLMs for MT, which we summarize briefly for completeness. A number of papers \cite{agrawal2022context,zhang2023prompting,jiao2023chatgpt,hendy2023good} investigate prompt quality and source proximity using methods similar to ours but with different LLMs, notably GPT-3.5, GPT-4 and their instruction-tuned counterparts. Their findings are in line with ours, with the exception of \citet{agrawal2022context}, who achieve significant gains using lexical matching augmented with a diversity mechanism to select prompts. Apart from differences in model and setting, a potentially salient discrepancy is their emphasis on BLEU rather than neural metrics to measure performance. Other interesting work that conditions prompts on source segments uses dictionaries to supply translations in low-resource settings \cite{ghazvininejad2023dictionary,lu2023chain}, or chain-of-thought inspired prompts that elicit keywords, topic, and related examples from the model itself \cite{he2023exploring}.

Further recent work looks at the role of data, attributing LLM MT capabilities to the presence of incidental bilingual examples \cite{briakou2023searching}, or showing that parallel data \cite{schioppa2023cross}, dictionaries \cite{jones2023bilex}, or restriction to bilingual settings \cite{garcia2023unreasonable} can boost performance in smaller LMs. 
Another popular line aims at controlling various properties of translations such as formality or use of specified terminology, either statically \cite{garcia2023unreasonable,moslem2023adaptive} or with human interaction \cite{pilault2023interactive}.
Finally, there is extensive work on analyzing the translation output of LLMs, generally finding that it is more fluent than accurate \cite{hendy2023good, anonymous2023does}, good at handling document context \cite{wang2023document,karpinska2023large}
but also prone to problems such as hallucination \cite{zhang2023prompting,guerreiro2023hallucinations},
and frequently sub-par in low-resource settings \cite{zhu2023multilingual,bawden2023investigating}

\section{Prompting for Machine Translation}
\label{sec:method}

For a general task, prompting an LLM to generate a desired output $y$ from an input $x$ can involve many steps \cite{liu2021pre}, including template generation, slot filling, answer search, and answer mapping.
In MT, the answer search and mapping processes are simplified because the answers generated by the LLM can be used directly; we simplify further by using a fixed template.
What we explore in depth is the slot filling portion; in particular, we test a variety of methods to select few-shot examples for the prompt.

In initial experiments we determined that for few-shot prompting the exact form of the template is unimportant, see Appendix~\ref{sec:prompt-exploration} for details.
Following this observation, we decided to adopt simple templates where each example if preprended by the corresponding language name.
These results in prompts of the form (for $n$-shot prompting):
\newcommand{\niceBrackets}[1]{&\texttt{[}&\omit\hfill $#1$\hfill\texttt{]}}
\newcommand{\tightX}{\hspace{-4.7em}}
\newcommand{\tightY}{\vspace{-0.25em}}
\begin{alignat*}{5}
   \tightX &\texttt{[source]: }& \niceBrackets{X_1} \tightY \\
   \tightX &\texttt{[target]: }& \niceBrackets{Y_1} \tightY \\
   \tightX &\texttt{...} \tightY \\
   \tightX &\texttt{[source]: }& \niceBrackets{X_n} \tightY \\
   \tightX &\texttt{[target]: }& \niceBrackets{Y_n} \tightY \\
   \tightX &\texttt{[source]: }& \niceBrackets{X} \tightY \\
   \tightX &\texttt{[target]: } \tightY
\end{alignat*}
where \texttt{[source]} and \texttt{[target]} are instantiated with the names in English of the source and target languages, e.g.\ \texttt{English} and \texttt{German}.
Note that this scheme has been found to be present in the training data as a marker for multilingual content \cite{briakou2023searching}.
Each slot pair $(X_i, Y_i)$ is filled with a translation example for these languages, and the final slot $X$ is filled with the current source text. Our algorithm for $n$-shot translation from a source text $x$ to a target text $y$ is:
\begin{enumerate}
    \item Choose translation example pairs $(x_1,y_1)$ ...  $(x_n,y_n)$. In general, these can depend on $x$.
    \item Plug the example pairs and $x$ into the template. Condition \palm on the resulting string.
    \item Perform a greedy search,\footnote{We found that using a sampling temperature other than 0 tended to degrade translation quality.} stopping when the model outputs a newline.
    \item Output the predicted suffix verbatim as $y$.
\end{enumerate}

Example selection operates in two phases: first choose a pool containing parallel text, then choose examples from the pool. Choosing the pool lets us control global attributes of examples such as domain and average quality. Our baseline method for choosing examples is to select them randomly from the pool. We also experiment with selecting examples that are ``closest’’ to the source text, on the hypothesis that such examples will help guide the model to produce similar translations.

To find relevant examples, we use $k$-nearest neighbor~(\kNN) search on the source side of our parallel pool, inspired by \citet{khandelwal2021nearest}.
We carry out the search itself using the method of \citet{avq_2020}\footnote{Available at \url{https://github.com/google-research/google-research/tree/master/scann}.}, and investigate two possible representations of the sentences, with associated distance measures:

\textbf{Bag-of-words (BOW):} Each sentence is represented by a (sparse) vector of counts associated with words in the vocabulary.
    As the associated distance measure we use cosine distance.
    This representation focuses on the surface form of the words, and thus favors lexical similarity between the examples.
    
\textbf{\roberta:} Sentences are represented as embeddings in the space defined by \roberta \cite{liu2019roberta}, a multilingual transformer-based model, with Euclidean distance used for retrieval. We expect these embeddings to reflect the semantics of the sentence, and thus retrieve prompts that are relevant to their subject matter.\footnote{Note that it would be conceivable to use \palm itself as embedding model, which would provide a representation (and associated similarity measure) closer to the application that we are targeting.
However, due to the high computational cost and large amounts of data (for some experiments we embed the totality of the WMT training data) we decided to use a smaller model.}

\section{Data}

\begin{table}
    \centering
    \begin{tabular}{llrr}
        \toprule
        LP & Year & \#sents & Ref \\
        \midrule
         \langdir{en}{de} & 2021 & 1002 & C \\
         \langdir{de}{en} & 2021 & 1000 & B \\
         \midrule
         \langdir{en}{zh} & 2021 & 1002 & A \\
         \langdir{zh}{en} & 2021 & 1948 & A \\
         \midrule
         \langdir{en}{fr} & 2014 & 3003 & N/A \\
         \langdir{fr}{en} & 2014 & 3003 & N/A \\
         \bottomrule
    \end{tabular}
    \caption{
    Test set information, including the newstest dataset year and, when applicable, the reference we use for scoring.
    }
    \label{tab:test_sets}
\end{table}

\begin{table}
    \centering
    \begin{tabular}{llrr}
        \toprule
        \multirow{2}[2]{*}{LP} & \multirow{2}[2]{*}{Pool} & \multicolumn{2}{c}{Size} \\
        \cmidrule(lr){3-4} && \langdir{en}{X} & \langdir{X}{en} \\
        \midrule
        \multirow{3}{*}{\langbidir{de}{en}}
        & \wmtfull     & \multicolumn{2}{c}{96M} \\
        & \wmtdev       & 11\,732   & 13\,060     \\
        & \highend     &   \multicolumn{2}{c}{152 para.}  \\
        \midrule   
        \multirow{3}{*}{\langbidir{zh}{en}}   
        & WMT-full     & \multicolumn{2}{c}{55M} \\
        & \wmtdev       & 7\,481    & 5\,916      \\
        & \highend     &  \multicolumn{2}{c}{170 para.}  \\
        \midrule   
        \multirow{3}{*}{\langbidir{fr}{en}}   
        & WMT-full     & \multicolumn{2}{c}{40M} \\
        & \wmtdev       & 2\,886    & 2\,957      \\
        & \highend     &  \multicolumn{2}{c}{\hspace{0.6em}98 para.}  \\
        \bottomrule
    \end{tabular}
    \caption{
    Size of the different prompt pools, measured in sentences for the WMT sets and in paragraphs for the \highend pool.
    }
    \label{tab:pool_sizes}
\end{table}

We experiment with translation into and out of English for Chinese, French and German.
After English (78.0\%), German (3.5\%) and French (3.3\%) are the two largest languages in \palm's 780B token training corpus; Chinese (0.4\%) is the 15\/th largest, and it also represents an inherently more difficult translation task.
To facilitate comparisons with recent SOTA systems, and to minimize the chance of overlap with \palm's training corpus, we test on news data from the WMT 2021 evaluation campaign \cite{akhbardeh-etal-2021-findings}.
Since French was not included in WMT21, we use data from WMT14; apart from being older, these test sets are not purely source-original~\cite{freitag-etal-2019-ape} like the more recent ones.
Table~\ref{tab:test_sets} shows statistics for our test data.

For prompt selection, we use three distinct pools: the full WMT training corpus for each language pair (\wmtfull), the corresponding WMT development sets (\wmtdev), and a manually-curated ``high-end’’ pool.
Sizes are shown in Table~\ref{tab:pool_sizes}.
The \wmtfull pool is largest and offers the highest probability of close \kNN matches, but it is crawled text drawn from sources of varying quality. The \wmtdev pool has generally better quality, and is a closer domain match to our test set; to encourage \palm to produce more natural text, we included only target-original texts.\footnote{As identified by \sacrebleu.} 
For \langbidir{German}{English} and \langbidir{Chinese}{English} we include all the news test sets from 2010 to 2020.
As \langbidir{English}{French} was discontinued after 2015, we used sets from 2010 to 2013, augmented with newsdiscussion2015.

The \highend pool comes from websites containing bilingual articles that we judged to be professionally edited, with native or near-native quality in both languages. The articles are drawn from various domains (biography, business, commentary, culture, fashion, food, news, and obituary), with the news domain of the test sets comprising less than 50\% for each language. We treat these articles as symmetrical, and use them as prompt sources in both translation directions. Due to the non-literal nature of the translations, there is frequently no 1-1 correspondence between sentence pairs, so we extract aligned paragraphs for prompting.
More detailed information about the \highend pool is provided in Appendix~\ref{sec:pool-details}.

\section{Experiments}
\label{sec:experiments}

\todo{Include experiments that show explicit degradation in output quality as prompt quality declines?}

For compatibility with \citet{chowdhery2022palm}, we ran all experiments at the sentence level, translating each test sentence individually and in isolation from its context. This deprives \palm of the ability to exploit the longer contexts it was exposed to during training, but it matches the operating mode of our baselines (including SOTA baselines), and facilitates evaluation.\footnote{Evaluation of document-level translations is complicated by potentially non 1-1 sentence correspondences, resulting in long translation units that are truncated by \bleurt and can be difficult for humans to rate reliably.}
We leave an exploration of potential gains from conditioning on longer histories to future work.

In preliminary experiments, we varied the number of shots from 0 to 10, and found clear performance gains as we increased the number of shots, with diminishing returns after 5 sentence pairs (see Appendix~\ref{sec:prompt-exploration}). Accordingly we report all results on the WMT pools in the 5-shot setting, where each shot is a single sentence pair, matching the configuration in \citet{chowdhery2022palm}. For the \highend pool, lacking 1-1 sentence alignments, we use 1-shot examples, where each shot is a single paragraph pair. This provides roughly the same quantity of text as 5-shot with sentences, although it creates a stylistic mismatch with our test setup, as we still translate on a sentence-by-sentene basis, as in the other conditions.

When randomly selecting examples, we observed that there is little variability in automatic scores when selecting different samples\footnote{Note that this holds for document level scores. The effect on single sentences can still be very important, cf.\ Figure~\ref{fig:random1_vs_random2}.} (see Appendix~\ref{sec:random-runs}).
For the results reported in this section, we let \palm produce translations with 5 different seeds and we selected the run with the median \bleurt score.
Translation time was some orders of magnitude longer than a dedicated translation system.

\todo[Any]{Create a table and supporting text so that reporting the variance of random selection strategies on different pools becomes a mini-contribution of this work. $\rightarrow$ Expand Appendix~\ref{sec:random-runs}.}

Following recent recommendations \cite{kocmi-etal-2021-ship, freitag-etal-2021-experts} we favour neural metrics (\bleurt in our case) over \bleu, although we also report \bleu scores for completeness.
We use a cased version of \bleurt{}~\cite{sellam-etal-2020-bleurt} that is based on Rem\textsc{bert} \cite{chung2020rethinking}.
We use \bleu as implemented in \sacrebleu \footnote{\sacrebleu signature: {\tt nrefs:1|case:mixed|eff:no| tok:}TOK{\tt|smooth:exp|version:2.1.0}, where TOK is \texttt{13a} or \texttt{zh}.} \cite{post-2018-call}, with \texttt{zh} tokenization for English-Chinese, and \texttt{13a} tokenization for all other languages.

To perform human evaluation, we hired professional translators (7 for En$\to$De, 5 for De$\to$En, 4 for Zh$\to$En, and 4 for En$\to$Zh) and measure translation quality with a document-context version of \mqm \cite{lommel2014multidimensional} which mimics the setup proposed in \citet{freitag-etal-2021-experts}.
This includes using the same error categories, severity levels and error weighting schema.
As suggested in the study, we weight each major error with~$5$ and each minor error with~$1$, except for minor punctuation errors which get a score of~$0.1$.
We depart from \citet{freitag-etal-2021-experts} in using only a single annotator per segment, and in not imposing a limit of 5 errors per sentence.
Additionally, due to technical restrictions on the length of an evaluation session, we limited the \mqm
evaluation to the first 12 segments per document.


\subsection{Selection strategies and pools}
\label{sec:selection_and_pools}

We warm up by comparing example selection strategies on the two WMT pools, using automatic metrics to evaluate quality on \langbidirtight{English}{German}. Results are shown in Table~\ref{tab:selection_and_pools}.
The main observation is that the choice of pool is much more important than the selection method: the results for \wmtdev are notably higher than those for \wmtfull across all settings. When comparing \kNN selection methods, \roberta is more effective than BOW, but it does not provide a consistent advantage over random selection. 

We conjecture that the quality of an example is more important than its proximity to the current source sentence. The larger size of the full WMT pool means that the \kNN approaches will in general be able to find examples that are closer to each source sentence than those from the dev pool, but any resulting gain is offset by the greater risk that an example from the full pool will be a poor translation (since we match only on the source side). Interestingly, had we relied only on \bleu, we would have concluded that the choice of pool is unimportant, and that random selection consistently outperforms \kNN.

\begin{table}
    \centering
    \begin{tabular}{lllrHr}  
    \toprule
    LP & Pool & Selection & \makebox[6ex]{\bleurt} & \comet & \bleu \\
    
    \midrule
    
    \multirow{5}[3]{*}{\rotatebox{90}{\langdir{en}{de}}}
        & \multirow{3}{*}{full}
        & random         & 71.8 & 49.6 & 32.9 \\
        && \kNN BOW      & 71.7 & 47.6 & 32.4 \\
        && \kNN \roberta & 73.0 & 52.1 & 32.5 \\
        \cmidrule(lr){3-6} 
        & \multirow{2}{*}{dev}
        & random        & 74.8 & 56.6 & 32.8 \\
        && \kNN \roberta & 74.8 & 56.4 & 32.3 \\ 
         
    \midrule
    
    \multirow{5}[3]{*}{\rotatebox{90}{\langdir{de}{en}}}
        & \multirow{3}{*}{full}
        & random         & 74.8 & 60.6 & 38.4 \\
        && \kNN BOW      & 72.7 & 53.7 & 36.9 \\
        && \kNN \roberta & 73.8 & 56.5 & 35.4 \\
        \cmidrule(lr){3-6}
        & \multirow{2}{*}{dev}
        & random         & 75.9 & 63.7 & 38.0 \\
        && \kNN \roberta & 75.8 & 62.7 & 37.2 \\
        
    \bottomrule
        
    \end{tabular}
    \caption{
    Comparison of example selection strategies on the \wmtfull and \wmtdev pools. Values for random selection are averaged over 5 runs.
    }
    \label{tab:selection_and_pools}
\end{table}

\subsection{Results on all language pairs}
\label{sec:full-results}

\begin{table*}
    \newcommand{\best}[1]{$\textbf{#1}$}
    \newcommand{\bestSign}[1]{\best{#1}\rlap{$^{\dagger}$}}
    
    \begin{subtable}{\textwidth}
    \centering
    \begin{tabular}{lllccHc}  
        \toprule
        LP    & \multicolumn{2}{l}{System}  & \mqm$\downarrow$ & \bleurt$\uparrow$ & \comet$\uparrow$ & \bleu$\uparrow$  \\
        \midrule                                                      
                                                                      
        \multirow{6}{*}{\langdir{en}{de}}                             
              & \multicolumn{2}{l}{WMT21 Facebook Submission~\cite{tran-etal-2021-facebook}} & \bestSign{1.18} & \best{76.9}    & \best{63.1}    & \best{42.0}  \\
              & \gtransshort               && 1.59 & 75.7    & 60.7    & 39.8  \\
              \cmidrule(lr){2-7}                                            
              & \multirow{4}{*}{\palm}                                  
                      & \wmtfull random              & 1.90 & 73.7    & 53.0    & \best{32.9}  \\
              &       & \wmtfull \kNN                & 1.93 & 73.0    & 52.1    & 32.5  \\
              &       & \wmtdev random      & \best{1.58} & \best{74.8}    & \best{57.0}    & 32.8  \\
              &       & \highend random     & 1.67           & 74.7    & 56.6    & 32.9  \\
                                                                        
        \midrule                                                        
                                                                        
        \multirow{6}{*}{\langdir{de}{en}}                               
              & \multicolumn{2}{l}{WMT21 Facebook Submission~\cite{tran-etal-2021-facebook}} & \bestSign{1.31} & \best{76.9}    & \best{67.0}    & \best{41.9}  \\
              & \gtransshort               && 1.71 & 76.4    & 66.0    & 40.9  \\
              \cmidrule(lr){2-7}                                            
              & \multirow{4}{*}{\palm}                                  
                      & \wmtfull random              & 2.38 & 74.7    & 60.2    & 38.3  \\
              &       & \wmtfull \kNN                & 3.03 & 73.8    & 56.5    & 35.4  \\
              &       & \wmtdev random      & 1.92 & \best{75.9}    & \best{63.8}    & 38.0  \\
              &       & \highend random     & \best{1.89} & 75.8    & 63.2    & \best{38.8}  \\
       \bottomrule
    \end{tabular}
    
    \todo[Viresh]{\mqm update ZhEn \mqm results}
    
    \subcaption{\langdirtight{German}{English} (nt2021). All \mqm results labelled with $\dagger$ are significantly better than all other systems based on PERM-BOTH pair-wise significance testing~\cite{koehn-2004-statistical} with $p=0.05$.}
    \label{tab:results_de-en}
    \end{subtable}
    
    \vspace{1em}
    
    \begin{subtable}{\textwidth}
    \centering
    \begin{tabular}{lllccHc}  
        \toprule
        LP    & \multicolumn{2}{l}{System}  & \mqm$\downarrow$ & \bleurt$\uparrow$ & \comet$\uparrow$ & \bleu$\uparrow$  \\
        \midrule                                                      
                                                                      
        \multirow{6}{*}{\langdir{en}{zh}}                             
              & \multicolumn{2}{l}{WMT21 WeChat Submission~\cite{zeng-etal-2021-wechat}} & \bestSign{2.47} & \best{66.6}    & \best{50.1}    & \best{36.9}  \\
              & \gtransshort               && 3.23 & 65.0    & 46.5    & 36.2  \\
              \cmidrule(lr){2-7}                                            
              & \multirow{4}{*}{\palm}                                  
                      & \wmtfull random              & 4.35 & 62.2    & 34.5    & 28.6  \\
              &       & \wmtfull \kNN                & 5.06 & 60.7    & 29.0    & 28.5  \\
              &       & \wmtdev random      & \best{3.24} & \best{64.1}    & \best{34.5}    & 29.2  \\
              &       & \highend random     & 3.70 & 63.9    & 41.8    & \best{29.6}  \\
                                                                        
        \midrule                                                        
                                                                        
        \multirow{6}{*}{\langdir{zh}{en}}                               
              & \multicolumn{2}{l}{WMT21 Borderline Submission~\cite{wang-etal-2021-tencent}} &   \best{3.11}   & \best{70.0}    & \best{48.4}    & \best{33.4}  \\
              & \gtransshort               &&  3.12    & 69.5    & 47.2    & 32.2  \\
              \cmidrule(lr){2-7}                                            
              & \multirow{4}{*}{\palm}                                  
                      & \wmtfull random     &   3.95   & 67.2    & 39.1    & \best{25.8}  \\
              &       & \wmtfull \kNN       &   4.06    & 65.8    & 33.3    & 23.8  \\
              &       & \wmtdev random      &  \best{3.60}    & 67.5    & \best{40.2}    & 25.3  \\
              &       & \highend random     &  3.89    & \best{67.7}    & 40.0    & 25.1  \\
              
       \bottomrule
    \end{tabular}
    
    \subcaption{\langdirtight{Chinese}{English} (nt2021). All \mqm results labelled with $\dagger$ are significantly better than all other systems based on PERM-BOTH pair-wise significance testing~\cite{koehn-2004-statistical} with p=0.05.}
    \label{tab:results_zh-en}
    \end{subtable}
    
    \vspace{1em}
    
    \begin{subtable}{\textwidth}
    \centering
    \begin{tabular}{lllcHc}  
        \toprule
        LP    & \multicolumn{2}{l}{System}  & \bleurt$\uparrow$ & \comet$\uparrow$ & \bleu$\uparrow$  \\
        \midrule                                                
                                                                
        \multirow{5}{*}{\langdir{en}{fr}}                       
              & \gtransshort                && 76.5   & ????    & 45.7 \\
              \cmidrule(lr){2-6}                                     
              & \multirow{4}{*}{\palm}                           
                    & \wmtfull random                & \best{75.9}    & \best{75.7}    & \best{42.3}  \\
              &     & \wmtfull \kNN                  & 75.3    & 74.6    & 41.8  \\
              &     & \wmtdev random        & 75.4    & 75.3    & 41.9  \\
              &     & \highend random       & 75.2        &         & 38.6      \\
                                                                 
        \midrule                                                 
                                                                 
        \multirow{5}{*}{\langdir{fr}{en}}                        
              & \gtransshort               && 77.7    & 73.1    & 43.2  \\
              \cmidrule(lr){2-6}                                     
              & \multirow{4}{*}{\palm}                           
                      & \wmtfull random              & \best{77.7}    & \best{72.6}    & \best{42.7}  \\
              &       & \wmtfull \kNN                & 77.3    & 71.6    & 41.2  \\
              &       & \wmtdev random      & 77.2    & 71.8    & 42.1  \\  
              &       & \highend random     & 77.6    &         & 40.4  \\
              
       \bottomrule
    \end{tabular}
        
    \subcaption{\langdirtight{French}{English} (nt2014).}
    \label{tab:results_fr-en}
    \end{subtable}
    
    \caption{Translation results for all language pairs. Values for random selection are the \bleurt median of 5 runs.}
    \label{tab:results}
    
    \vspace{-0.5ex}  
\end{table*}

Table~\ref{tab:results} contains our main results, 
for \langbidir{German}{English},
\langbidir{Chinese}{English}, and \langbidir{French}{English}. For each language pair, we ran \palm with random selection on all three pools and with \kNN \roberta on the \wmtfull pool. We compared these systems to output from the best performing system in the 2021 WMT evaluation campaign for German and Chinese, and for off-the-shelf \gtrans for all six language pairs. We evaluate with \bleu and \bleurt as in the previous section, augmented with human \mqm assessments for German and Chinese. French is a special case, as its evaluation set is eight years old, and it is difficult to ensure that any of the MT systems we evaluate have not been exposed to it during training. We include it mostly for the purposes of comparison to \citet{chowdhery2022palm}, and do not provide SOTA results or perform human evaluation. 

Comparing \palm results for German and Chinese, the pattern from the previous section holds up: random selection from the \wmtdev pool outperforms selection from the full pool. \mqm scores correlate well with \bleurt for these results. Despite domain and style mismatch, results for the \highend pool are very similar to those for \wmtdev---closer than any results on the full pool---adding support to the hypothesis that example quality is the main determinant of \palm's output quality.

The French results reverse  the general pattern. For this language pair, random selection from the \wmtfull pool does best, although the results for all methods are fairly similar, with a difference of approximately 0.5 \bleurt between the best and worst. One potential explanation is the age and quality of newstest2014, as WMT test-set creation has dramatically improved since then.

Turning to a comparison between \palm and conventional MT systems, the specialized SOTA systems have a substantial advantage of between 1 and 3 \bleurt points over the best \palm results, a gap that is reflected in their much lower \mqm scores. The difference is narrower for the general-purpose \gtrans system: less than 1 \bleurt except for \langdirtight{Chinese}{English} (1.8), with \langdirtight{French}{English} at parity. \palm's performance relative to the best MT system for each language pair
is generally better when translating into English, where it is lower by 1.0, 2.3, and 0.0 \bleurt for German, Chinese, and French, compared to drops of 2.1, 2.5, and 0.6 in the reverse direction. 

\begin{table}
\centering
\begin{tabular}{lllrr}
\toprule
LP & Sev. & Category & \palm & SOTA \\
\midrule
\langdir{de}{en} & Major & Omission & 51 & 19 \\
\langdir{en}{de} & Major & Omission & 26 & 7 \\
\langdir{zh}{en} & Major & Omission & 109 & 42 \\
\langdir{en}{zh} & Major & Omission & 80 & 46 \\
\langdir{de}{en} & Minor & Awkward & 73 & 81 \\
\langdir{en}{de} & Minor & Awkward & 166 & 144 \\
\langdir{zh}{en} & Minor & Awkward & 205 & 284 \\
\langdir{en}{zh} & Minor & Awkward & 115 & 142 \\
\bottomrule
\end{tabular}
\caption{Selected \mqm error count comparisons between \palm \wmtdev random and SOTA. Omission is a subcategory of Accuracy errors, and Awkward is a subcategory of Style.
Full details are provided in Appendix~\ref{sec:detailed-mqm-scores}.}
\label{tab:mqm_cats}
\end{table}

The \mqm results show some interesting characteristics of translations produced by \palm.
In all language pairs evaluated, fluency \mqm scores for \palm are generally similar to those for SOTA systems, while accuracy scores are lower.
The accuracy gap is dominated by Major Accuracy/Omission errors, followed by inconsistent patterns of other Accuracy/* errors across language pairs.
In some languages, the best-performing \palm systems make fewer Style/Awkward errors than SOTA. Table~\ref{tab:mqm_cats} shows a selection of \mqm error counts for  \palm \wmtdev random and SOTA systems; full details are provided in Appendix~\ref{sec:detailed-mqm-scores}.

\subsection{Comparison to previous results}

Our only results that are directly comparable to the few-shot results from \citet{chowdhery2022palm} are the \wmtfull \bleu scores in table~\ref{tab:results_fr-en} (WMT14 French test-set). Our result for \langdirtight{French}{English} matches theirs exactly, but our score for \langdirtight{English}{French} is lower by 1.7 (42.3 versus 44.0). We attribute this discrepancy to their use of the \sacrebleu \texttt{intl} tokenizer; when we evaluate our output using this version, we obtain matching scores.

Our general finding that \palm's into-English performance is better than the reverse direction matches the conclusion from \citet{chowdhery2022palm}, while our comparison with recent SOTA systems on current test sets 
contrasts with their results indicating that \palm can rival supervised performance in older settings.

\section{Analysis}

In this section we delve further into various aspects of \palm's MT performance.

\subsection{\kNN versus random prompts}
\label{sec:knn_vs_random}

To understand the performance difference between \kNN \roberta and randomly-selected examples, 
we performed a qualitative analysis, choosing sentences with the largest \bleurt difference between the two systems.
Table~\ref{tab:example_kNNWins} in Appendix~\ref{sec:example-prompts} shows an example where the \kNN system correctly retrieves relevant translation examples in the football domain, guiding \palm to produce a better translation
than the random selection system. 
This contrasts with the example in Table~\ref{tab:example_randomWins}, where the retrieved source sentences are also from the relevant domain, but all have alignment errors, causing \palm to generate hallucinated output.
In general, random selection is also prone to landing on alignment errors, but as each prompt is selected independently, the odds that \emph{all} examples will be errors are low. 
An informal analysis of \kNN examples indicates that if one non-parallel prompt is selected, the others also tend to be of poor quality, perhaps due to corpus alignment errors that are concentrated in particular documents or topics. Since \kNN matches only on the source side, it is not robust to this noise.

\subsection{Example Translations}
\label{sec:example_translations}

Example translations comparing \palm and SOTA systems for \langdirtight{German}{English} and \langdirtight{English}{Chinese} are given in Appendix~\ref{sec:example_translations}, in Table~\ref{tab:deen_examples} and Table~\ref{tab:enzh_examples}, respectively.
We compared the translations of both systems and chose examples that are short, but include the most frequent patterns that we observed also in longer translations. In general, \palm's translations are less literal when compared to supervised NMT systems. Even though this is one of the strengths of \palm, it occasionally misses some important information in the source or hallucinates facts not present in the source sentence. The supervised models on the other hand are faithful to the source; this reduces the risk of omission and addition errors, but occasionally leads to translations that are not natural in the target language (e.g. translating street names or using the wrong time format). These findings are in line with the \mqm results presented in section~\ref{sec:full-results}.


\subsection{Overlap of test and training data}

\begin{table}
    \centering
    \begin{tabular}{llr}
    \toprule
    Year & LP & \% Clean \\ \midrule
    \multirow{2}{*}{2014} & \langdir{fr}{en} & \textbf{69.2} \\
         & \langdir{en}{fr} & 93.6 \\ \midrule
    \multirow{2}{*}{2016} & \langdir{de}{en} & \textbf{80.3} \\ 
         & \langdir{en}{de} & 97.3 \\ \midrule
    \multirow{4}{*}{2021} & \langdir{en}{de} & 99.6 \\
         & \langdir{en}{zh} & 99.7 \\
         & \langdir{de}{en} & 97.9 \\
         & \langdir{zh}{en} & 98.1 \\
    \bottomrule
    \end{tabular}
    \caption{The size of clean (lacking 15-gram target-side overlap with \palm training data) versions of test sets for various WMT years and language pairs \label{tab:overlap_percent}}
\end{table}

One major change with respect to \citet{chowdhery2022palm} is our use of more recent WMT test sets, 
which are unlikely to overlap with \palm's training data.\footnote{Here we measure target-side overlap only;
we assume there is no substantial parallel data in \palm's training corpus, and therefore no substantial parallel overlap.} We test this hypothesis using the technique from \citet{chowdhery2022palm}, which involves measuring high-order $n$-gram matches; specifically, we measure 15-gram overlap as tokenized by the mBERT tokenizer~\cite{devlin-etal-2019-bert}.\footnote{We selected the mBERT tokenizer, as opposed to the \palm's sentence-piece tokenizer, because it decouples the measurement of overlap from the model under test.}
For test sequences with fewer than 15 tokens, we consider them overlapping if the complete sequence is found as a subsequence of a training example. We report the degree of overlap by showing the percentage of original
test examples that survive in the clean test set after removing overlap in Table~\ref{tab:overlap_percent}.
This confirms that the older \langdirtight{French}{English} and \langdirtight{German}{English} sets have substantial overlap with \palm's training data,
while the newer test sets, whether into or out of English, have much smaller overlapping portions.

\citet{chowdhery2022palm} also measure the effect of test-set overlap on translation quality, comparing scores on the original test set to the clean set with overlapping examples removed.
In section~\ref{sec:overlap_analysis} we report similar scores for the older test sets, and extend 
the analysis to calibrate the effect of overlap on MT evaluation, by comparing to an overlap-free off-the-shelf system.



\section{Conclusion}

We perform a careful assessment of the sentence-level MT capabilities of \palm, which we compare to SOTA and a current off the shelf (COTS) MT system for three high-resource languages---German, Chinese, and French---into and out of English, using the latest test sets from WMT.
We chose to focus on a small set of high-resource language pairs in order to test the claims of the original PaLM paper, which are most striking for these pairs.
The time and expense of performing high-quality human evaluations precluded a broader investigation.

Comparing \kNN and random strategies for selecting 5-shot translation examples to instantiate fixed prompt templates, we find that {\kNN}'s potential advantage in identifying examples relevant to the source sentence is outweighed by its susceptibility to corpus noise. Choosing examples randomly from small, high-quality pools works well, and performance appears to be independent of the domain and translation style of the pool, suggesting that example quality is the most important factor. 

Using both the \bleurt metric and \mqm human evaluations, we show that {\palm}'s performance, while very impressive for a system never deliberately exposed to parallel text, still significantly lags that of competition-grade SOTA systems on recent WMT test sets, and to a lesser extent the performance of COTS systems as well. This contrasts with some of the findings of \citet{chowdhery2022palm}. As in that work, we find that performance into English is somewhat better than the reverse direction. Finally, we perform an extensive analysis of the characteristics of {\palm}’s MT output, notably finding that in all languages we tested it tends to be creative and fluent but prone to omissions and other accuracy errors; broadly speaking, it
matches the fluency but lags the accuracy of conventional NMT.

In future work we look forward to testing \palm on document-level translation tasks, unleashing its formidable capacity for leveraging long contexts. We would also like to explore prompt tuning methods that are more sophisticated than the hard-prompt setting we adopted for this paper, particularly to see if these might offer a way to tighten up {\palm}’s MT accuracy without destroying its impressive ability to generate highly-fluent text.

\section*{Limitations}

As we use only a small number of language pairs, it is not clear how general our conclusions are; in particular, they pertain only to languages that are well represented in \palm's training corpus, and only to translation into and out of English. Our restriction to independent sentence-level translations may have caused us to underestimate \palm's true capabilities, since some of the accuracy problems we observed might be considered less severe in the context of whole-document translation where less literal translations are more typical. Our exploration of prompting barely scratches the surface of the many methods that have been proposed for adapting LLMs to particular tasks, and we may have missed a technique that produces higher-quality translations than we observed. Finally, the human evaluation we rely on to provide our most accurate results is necessarily subjective, and if we were to have carried out the evaluation with different raters and a different methodology, our conclusions might well have been different.

\section*{Ethical Considerations}

Working with large language models comes with many ethical concerns that are discussed in detail in \newcite{brown2020language} and \newcite{chowdhery2022palm}. 
There, MT is often one task of many, while we focus on the question of proper example selection for few-shot prompting of MT,
which adds a few specific concerns.
Our conclusion that prompt quality is  important could lead one to build a system with prompts drawn from a small set of trusted sources; indeed, our \highend set is one such example of this.
In such a scenario, this small source will have an outsized impact on the output of the translation system, and one must be careful to manage issues of attribution and intellectual property.
Furthermore, an editorial choice defining high-quality language can potentially reduce quality for groups and topics not typically discussed in this style~\cite{gururangan-etal-2022-whose}.
Finally, by highlighting the power of few-shot examples, one might be tempted to turn example selection over to the users of a system.
There, special steps must be taken to avoid exposing users to biased or toxic outputs, which may be triggered by unconstrained prompting~\cite{gehman-etal-2020-realtoxicityprompts,costa-jussa-etal-2022-toxicity}.


\bibliography{anthology,custom}
\bibliographystyle{acl_natbib}


\appendix

\section*{Appendices}

\section{Prompt Exploration}
\label{sec:prompt-exploration}

As preliminary experiments we tried different prompting templates:

\newcommand{\promptDefault}{Language\xspace}
\newcommand{\promptCodes}{Codes\xspace}
\newcommand{\promptHeader}{Header\xspace}
\newcommand{\promptTextual}{Textual\xspace}
\newcommand{\promptDeutsch}{Deutsch\xspace}
\newcommand{\promptNone}{None\xspace}

\begin{description}
\item[\promptDefault] This is the prompt template used in the paper (see Section~\ref{sec:method}).
    It prepends the examples with the corresponding language name in English.
\item[\promptCodes] Like ``\promptDefault'', but instead of full English names, two-letter languages codes are used (e.g.\ ``en'', ``de'').
\item[\promptHeader] Like ``\promptDefault'', but the header ``Translate following sentences:'' is added.
\item[\promptTextual] A textual request for translating a sentence: ``Translate $X_n$ from English into German: $Y_n$'', where $X_n$ and $Y_n$ are the translation examples, as in Section~\ref{sec:method}.
The source sentence $X$ is given with the same template, but without specifying any translation.
\item[\promptDeutsch] Like ``\promptDefault'', but the language names are given in German (``Englisch'', ``Deutsch'').
\item[\promptNone] No added text. Source and target examples are just input one after the other.
\end{description}

\begin{table}
    \centering
    \begin{tabular}{lrrrrr}
        \toprule
        & \multicolumn{5}{c}{\# shots} \\
        Prompt           &    0 &    1 &    2 &    5 &   10 \\
        \midrule
        \promptDefault   & 63.9 & 69.1 & 71.7 & 73.6 & 74.4 \\
        \promptCodes     & 59.0 & 68.5 & 71.2 & 73.4 & 74.1 \\
        \promptHeader    & 72.4 & 69.1 & 70.7 & 73.4 & 74.1 \\
        \promptTextual   & 36.9 & 67.5 & 71.8 & 73.0 & 73.7 \\
        \promptDeutsch   & 72.6 & 70.8 & 71.9 & 73.5 & 74.1 \\
        \promptNone      &  3.2 & 38.5 & 59.6 & 73.0 & 74.1 \\
         \bottomrule
    \end{tabular}
    \caption{\bleurt results with different prompt templates and number of prompts for the \langdir{English}{German} translation directions. The prompt examples were randomly selected. The median of 5 runs are shown.}
    \label{tab:promptTemplates}
\end{table}

As shown in Table~\ref{tab:promptTemplates}, the choice of a prompting strategy has a crucial impact when the number of shots is low, but the effect is reduced when we increase the number of examples shown.
The number of examples also has a significant impact on translation quality.
We chose to work with 5 examples, as there are diminishing returns when increasing the number of prompts, and choosing a higher number has additional practical implications (e.g. possibly exceeding the maximum input length).

\section{High-end pool}
\label{sec:pool-details}

\begin{table}
    \centering
    \begin{tabular}{lrrr}
        \toprule
\multirow{2}[2]{*}{Genre} & \multicolumn{3}{c}{Proportion} \\
\cmidrule(lr){2-4} & \langbidir{en}{de} & \langbidir{en}{fr} &
\langbidir{en}{zh} \\
\midrule
biography  & 31\% & 20\% & -- \\
business   & --   & --   & 15\% \\
commentary & 25\% & 10\% & 16\% \\
culture    & --   & 44\% & 14\% \\
fashion    & 16\% & --   & -- \\
food       & --   &  8\% & -- \\
news       &  4\% & 18\% & 43\% \\
obituary   & 24\% & --   & 13\% \\
\bottomrule
    \end{tabular}
    \caption{Genre distributions for the \highend pool.}
    \label{tab:high_end_genre}
\end{table}

\begin{table*}
\fontsize{8}{10}\selectfont
    \centering
    \begin{tabular}{lrrl}
    \toprule
    LP & paras & words & URL \\
    \midrule
    \multirow{7}{*}{\rotatebox{90}{\langbidir{en}{de}}}
   & 4  &  255 & www.deutschland.de/en/news/new-supercomputer-in-operation \\
   & 4  &  208 & www.deutschland.de/en/news/patents-germany-ranks-second \\
   & 11 &  609 & www.deutschland.de/en/news/syrian-swimmer-yusra-mardini-provides-message-of-hope-at-olympics \\
   & 24 & 1787 & www.zeit.de/kultur/2019-12/schoenheit-fotografie-aesthetik-rankin-mitch-epstein-roger-ballen-english \\
   & 28 & 2817 & www.zeit.de/kultur/2020-07/ \\
   &&& desinformation-peter-pomerantsev-social-media-regulation-democracy/komplettansicht \\
   & 60 & 2961 & www.zeit.de/politik/ausland/2020-11/ \\
   & & & polarization-us-elections-democrats-republicans-donald-trump-family-division-english \\
   & 21 & 2757 & www.zeit.de/politik/deutschland/2015-11/helmut-schmidt-obituary-english/komplettansicht \\
    \midrule
    \multirow{9}{*}{\rotatebox{90}{\langbidir{en}{zh}}}
   & 30 & 1323 & cn.nytimes.com/asia-pacific/20220509/taiwan-china-covid/dual \\
   & 31 & 1317 & cn.nytimes.com/china/20220427/brownface-barrack-okarma-1968-hong-kong/dual \\
   &  6 &  780 & cn.nytimes.com/china/20220401/china-cheng-lei-australia/dual \\ 
   & 13 &  609 & cn.nytimes.com/china/20220421/china-eastern-crash-report/dual \\
   & 23 & 1520 & cn.nytimes.com/china/20220412/china-russia-propaganda/dual \\
   & 22 & 1373 & cn.nytimes.com/business/20220621/china-housing-real-estate-economy/dual \\
   & 13 &  478 & cn.nytimes.com/china/20220415/shanghais-food-crisis-prompts-residents-in-beijing-to-stockpile-supplies/dual \\
   & 26 & 1202 & cn.nytimes.com/obits/20220418/peng-ming-min-dead \\
   &  6 &  843 & https://cn.nytimes.com/world/20220330/solomon-islands-china/dual \\
    \midrule
 \multirow{13}{*}{\rotatebox{90}{\langbidir{en}{fr}}}
  &  6 &  846 & france-amerique.com/a-france-of-many-colors \\
  & 10 & 1177 & france-amerique.com/alice-guy-cinema-forgotten-pioneer \\
  & 10 & 1237 & france-amerique.com/americanization-is-back-did-it-ever-go-away \\
  &  8 &  666 & france-amerique.com/a-propos-a-hard-hitting-french-american-podcast \\
  &  3 &  457 & france-amerique.com/camille-laurens-a-womans-life \\
  &  8 &  970 & france-amerique.com/football-and-soccer \\
  &  6 &  377 & france-amerique.com/france-united-states-naval-battle-and-diplomatic-crisis \\
  &  7 &  615 & france-amerique.com/jeanne-damas-all-the-women-in-her-city \\
  &  6 &  631 & france-amerique.com/guedelon-building-a-castle-by-hand \\
  & 11 &  811 & france-amerique.com/raphael-francois-culinary-director \\
  & 12 &  874 & france-amerique.com/thierry-mugler-provocateur \\
  &  7 &  934 & france-amerique.com/winds-of-change-over-democracy \\
  &  4 &  255 & www.deutschland.de/en/news/new-supercomputer-in-operation \\
    \bottomrule
    \end{tabular}
    \caption{Sizes and provenance for articles in the \highend prompt pool. The \emph{words} column contains the number of English words (whitespace-separated character sequences) in each article.  }
    \label{tab:high-end}
\end{table*}

Table~\ref{tab:high-end} describes the \highend pool. All listed articles were manually downloaded in June--August 2022, and semi-automatically divided into bilingual paragraphs. Our \highend pool consists of all paragraphs from all articles. The domain breakdown for each language pair is shown in Table~\ref{tab:high_end_genre}.

\section{Variability of Random Runs}
\label{sec:random-runs}

Table~\ref{tab:random-runs} shows the automatic scores for random runs for the \langdirtight{German}{English} language pair.
It can be observed that the range of scores is quite small, less than 0.5 \bleurt points for all language directions. 
For both directions, the use of \wmtdev, as opposed to \wmtfull, for the random pool reduces the observed range in \bleurt by at least~$0.1$.

\begin{table*}
    \small
    \centering
    \begin{tabular}{llrrrrrrrrrr}
    \toprule
       &           & \multicolumn{5}{c}{\bleurt} & \multicolumn{5}{c}{\bleu} \\
       \cmidrule(lr){3-7} \cmidrule(lr){8-12}
    LP & Pool & Run 1 & Run 2 & Run 3 & Run 4 & Run 5 & Run 1 & Run 2 & Run 3 & Run 4 & Run 5 \\
    
    \midrule
    
    \multirow{2}{*}{\langdir{en}{de}}
        & full & 71.9 & 71.9 & 71.6 & 71.8 & 71.9 & 32.4 & 32.8 & 32.1 & 32.9 & 32.9 \\
        & dev  & 74.7 & 74.7 & 74.7 & 74.9 & 74.8 & 32.7 & 32.6 & 32.6 & 32.6 & 32.8 \\
         
    \midrule
    
    \multirow{2}{*}{\langdir{de}{en}}
        & full & 74.8 & 75.0 & 74.8 & 74.5 & 74.7 & 38.4 & 38.5 & 38.2 & 38.0 & 38.3 \\
        & dev  & 75.9 & 75.9 & 76.0 & 75.7 & 75.9 & 38.0 & 38.0 & 38.0 & 38.3 & 38.2 \\
        
    \bottomrule
    \end{tabular}
    \caption{Results for random runs for the \langdirtight{German}{English} translation direction.}
    \label{tab:random-runs}
\end{table*}

\section{Detailed \mqm Scores}
\label{sec:detailed-mqm-scores}

Table~\ref{tab:mqm-detailed-scores} presents \mqm scores for \palm \wmtdev random and SOTA systems in the four language pairs evaluated, along with
the breakdown of the scores into their Accuracy and Fluency components. Table~\ref{tab:mqm-detailed-error-counts} presents detailed \mqm error counts for
\palm \wmtdev random and SOTA systems in \langdirtight{en}{de} and \langdirtight{de}{en}.

\begin{table*}
    \centering

    \begin{tabular}{lrrrl@{\hspace{0.7cm}}rrr}
       \toprule
            & \multicolumn{3}{c}{\palm} && \multicolumn{3}{c}{SOTA} \\
            & \mqm$\downarrow$ & Accuracy$\downarrow$ & Fluency$\downarrow$ && \mqm$\downarrow$ & Accuracy$\downarrow$ & Fluency$\downarrow$ \\
       \midrule

\langdir{en}{de}       & 1.58 & 1.12 & 0.46 && 1.18 & 0.81 & 0.37 \\
\langdir{en}{zh}       & 3.24 & 2.69 & 0.52 && 2.47 & 1.96 & 0.48 \\
\midrule
\langdir{de}{en}       & 1.92 & 1.43 & 0.48 && 1.31 & 0.88 & 0.43 \\
\langdir{zh}{en}       & 3.60 & 2.97 & 0.62 && 3.11 & 2.43 & 0.68 \\

\bottomrule
    \end{tabular}
    \caption{\mqm scores for \palm \wmtdev random and SOTA systems, split into Accuracy and Fluency. Accuracy scores include "Accuracy/*," "Terminology/*," and
    "Non-translation!" error categories. Fluency scores include "Fluency/*," "Style/*," and "Locale/*" categories. The "Other" error category
    is not included in Accuracy or Fluency scores.}
    \label{tab:mqm-detailed-scores}
\end{table*}

\begin{table*}
    \centering

    \begin{tabular}{lrrrrl@{\hspace{0.7cm}}rrrr}
       \toprule
            & \multicolumn{4}{c}{\langdir{en}{de}} && \multicolumn{4}{c}{\langdir{de}{en}} \\
            & \multicolumn{2}{c}{\palm} & \multicolumn{2}{c}{SOTA} && \multicolumn{2}{c}{\palm} & \multicolumn{2}{c}{SOTA} \\
            & Major & minor & Major & minor && Major & minor & Major & minor \\
       \toprule

Non-translation!    & 0  & 0 & 0 & 0 && 2 & 0 & 0 & 0 \\
\midrule
Acc/Mistrans.       & 103  & 89 & 79 & 67 && 73 & 41 & 61 & 49 \\
Acc/Omission        & 26  & 6 & 7 & 3 && 51 & 33 & 19 & 11 \\
Acc/Addition        & 1  & 6 & 3 & 1 && 10 & 2 & 0 & 3 \\
Acc/Untranslated    & 12 & 4 & 14 & 0 && 6 & 7 & 5 & 8 \\
\midrule
Ter/Inappr          & 0  & 7 & 0 & 7 && 17 & 21 & 12 & 15 \\
Ter/Incons          & 0  & 4 & 0 & 4 && 1 & 5 & 1 & 7 \\
\midrule
Fl/Grammar          & 0  & 133 & 0 & 100 && 18 & 41 & 5 & 38 \\
Fl/Register         & 0  & 2 & 0 & 3 && 0 & 0 & 0 & 0 \\
Fl/Inconsistency    & 0  & 2 & 0 & 5 && 0 & 2 & 0 & 2 \\
Fl/Punctuation      & 0  & 260 & 0 & 31 && 1 & 38 & 2 & 29 \\
Fl/Spelling         & 0  & 12 & 0 & 16 && 0 & 16 & 0 & 17 \\
Fl/Encoding         & 0  & 0 & 0 & 0 && 0 & 0 & 0 & 2 \\
\midrule
St/Awkward          & 0  & 166 & 0 & 144 && 13 & 73 & 16 & 81 \\
\midrule
Locale/Date         & 0  & 0 & 0 & 0 && 0 & 1 & 0 & 5 \\
Locale/Name         & 0  & 0 & 0 & 0 && 2 & 8 & 2 & 5 \\
Locale/Time         & 0  & 0 & 0 & 0 && 0 & 5 & 0 & 5 \\
\midrule
Source Error        & 0  & 0 & 0 & 0 && 1 & 0 & 0 & 1 \\
Other               & 0  & 0 & 1 & 0 && 0 & 3 & 0 & 3 \\
\midrule\midrule
Total Errors        & 142 & 673 & 102 & 362 && 189 & 296 & 123 & 281 \\
\bottomrule
    \end{tabular}
    \caption{\mqm error counts for \palm \wmtdev random and SOTA systems for \langdirtight{en}{de} and \langdirtight{de}{en}.
    Abbreviations are as follows: ``Acc'':~Accuracy, ``Fl'':~Fluency, ``St'':~Style, ``Ter'':~Terminology, ``Inappr'':~Inappropriate for context, ``Incons'':~Inconsistent.}
    \label{tab:mqm-detailed-error-counts}
\end{table*}

\section{Significance numbers}

We calculate pairwise significance numbers based on PERM-BOTH pair-wise significance testing~\cite{koehn-2004-statistical, deutsch-etal-2021-statistical}. Results can be seen in Table~\ref{tab:significance_numbers}.

\begin{table*}[t]
    \centering
    \begin{tabular}{l l r r r r r r }
    \toprule
     &  & \multirow{2}{*}{SOTA} & \multirow{2}{*}{GTrans.} & WMT-dev & high-end & \multicolumn{2}{c}{WMT-full} \\
      & &  & &  random & random & random & kNN \\ \midrule
\multirow{7}{*}{de$\to$en} & MQM & 1.31 & 1.71 & 1.92 & 1.89 & 2.38 & 3.03 \\ \cmidrule{2-8}
&	SOTA	&	       -       	&	\hl{0.0}	&	\hl{0.0}	&	\hl{0.0}	&	\hl{0.0}	&	\hl{0.0} \\
&	Google Trans.	&	       -       	&	       -       	&	0.073	&	0.124	&	\hl{0.0}	&	\hl{0.0} \\
&	WMT-dev random	&	       -       	&	       -       	&	       -       	&	0.588	&	\hl{0.001}	&	\hl{0.0} \\
&	high-end random	&	       -       	&	       -       	&	       -       	&	       -       	&	\hl{0.001}	&	\hl{0.0} \\
&	WMT-full random	&	       -       	&	       -       	&	       -       	&	       -       	&	       -       	&	\hl{0.001} \\
\bottomrule
\multirow{7}{*}{en$\to$de} & MQM & 1.18 & 1.59 & 1.58 & 1.67 & 1.90 & 1.93 \\ \cmidrule{2-8}
&	SOTA	&	       -       	&	\hl{0.0}	&	\hl{0.0}	&	\hl{0.0}	&	\hl{0.0}	&	\hl{0.0} \\
&	Google Trans.	&	       -       	&	       -       	&	0.512	&	0.225	&	\hl{0.003}	&	\hl{0.003} \\
&	WMT-dev random	&	       -       	&	       -       	&	       -       	&	0.175	&	\hl{0.001}	&	\hl{0.0} \\
&	high-end random	&	       -       	&	       -       	&	       -       	&	       -       	&	\hl{0.021}	&	\hl{0.01} \\
&	WMT-full random	&	       -       	&	       -       	&	       -       	&	       -       	&	       -       	&	0.372 \\
\bottomrule
\multirow{7}{*}{zh$\to$en} & MQM & 3.11 & 3.12 & 3.60 & 3.89 & 3.95 & 4.06 \\ \cmidrule{2-8}
&	SOTA	&	       -       	&	0.447	&	\hl{0.0}	&	\hl{0.0}	&	\hl{0.0}	&	\hl{0.0} \\
&	Google Trans.	&	       -       	&	       -       	&	\hl{0.002}	&	\hl{0.0}	&	\hl{0.0}	&	\hl{0.0} \\
&	WMT-dev random	&	       -       	&	       -       	&	       -       	&	\hl{0.022}	&	\hl{0.006}	&	\hl{0.003} \\
&	high-end random	&	       -       	&	       -       	&	       -       	&	       -       	&	0.343	&	0.168 \\
&	WMT-full random	&	       -       	&	       -       	&	       -       	&	       -       	&	       -       	&	0.281 \\
\bottomrule
\multirow{7}{*}{en$\to$zh} & MQM & 2.47 & 3.23 & 3.24 & 3.70 & 4.35 & 5.06 \\ \cmidrule{2-8}
&	SOTA	&	       -       	&	\hl{0.0}	&	\hl{0.0}	&	\hl{0.0}	&	\hl{0.0}	&	\hl{0.0} \\
&	Google Trans.	&	       -       	&	       -       	&	0.488	&	\hl{0.004}	&	\hl{0.0}	&	\hl{0.0} \\
&	WMT-dev random	&	       -       	&	       -       	&	       -       	&	\hl{0.002}	&	\hl{0.0}	&	\hl{0.0} \\
&	high-end random	&	       -       	&	       -       	&	       -       	&	       -       	&	\hl{0.0}	&	\hl{0.0} \\
&	WMT-full random	&	       -       	&	       -       	&	       -       	&	       -       	&	       -       	&	\hl{0.0} \\
\bottomrule
    \end{tabular}
    \caption{p-values based on PERM-BOTH pair-wise significance testing~\cite{deutsch-etal-2021-statistical}. We highlight all numbers with p<0.05.}
    \label{tab:significance_numbers}
\end{table*}

\section{Example Prompts}
\label{sec:example-prompts}

Tables~\ref{tab:example_kNNWins} and \ref{tab:example_randomWins} show prompt examples where \kNN and random selection do better, respectively, as described in section~\ref{sec:knn_vs_random}.

\begin{table*}
    \newcommand{\prompt}[2]{\textbf{German:} #1 \textbf{English:} #2}
    

    \begin{subtable}{\linewidth}
    \centering \small
    \begin{tabular}{llp{13.3cm}}
    \toprule
    & Source    & "Wir haben die Pflichtaufgaben mit Meisterschaft und Pokal einfach hervorragend gemeistert. \\
    & Reference & “Quite simply, we have excellently mastered the necessary tasks for the Championship and the Cup. \\
    \midrule
    
    \multirow{12}{*}{\rotatebox{90}{\kNN \roberta}}
    & Hyp & "We have simply mastered the tasks of the championship and the cup excellently. \\
    & Prompt 1 & \prompt{Mit einer verstärkten Mannschaft holte die Mannschaft das Double aus Meisterschaft und Pokal.}{The decision paid off as the team achieved a league and cup double.} \\
    & Prompt 2 & \prompt{Darüber hinaus haben wir uns wichtige Meisterschaftspunkte im Kampf um den Vizetitel gesichert.“}{We have furthermore secured some important championship points in the fight about the vice champion’s title.”} \\
    & Prompt 3 & \prompt{Bring deine Mannschaft durch alle Spiele der Europameisterschaft und gewinne den Pokal!}{Take your team all the way through the Euro Cup stages and lift the trophy!} \\
    & Prompt 4 & \prompt{So konnte er die französische Meisterschaft, den nationalen Pokal sowie den Supercup gewinnen.}{He helped the club to win the national championship and the Supercup.} \\
    & Prompt 5 & \prompt{Roter Stern gewinnt in jener Saison das Double von Meisterschaft und Pokal.}{Red Star won their sixth double in this season.} \\
    
    \midrule
    \multirow{13}{*}{\rotatebox{90}{Random}}
    & Hyp & "We have the compulsory tasks with mastery and cup simply superbly mastered. \\
    & Prompt 1 & \prompt{Im November letzten Jahres war ein einzelner Steinadler und auch Bussarde im Blick der Kamera.}{In November last year a solitary golden eagle and buzzards too appeared in web camera view.} \\
    & Prompt 2 & \prompt{Teile: Modesto-14th Street, Stanislaus, California, Kalifornien-Luft-Qualitätsindex}{Share: Modesto-14th Street, Stanislaus, California, California Air Quality Index} \\
    & Prompt 3 & \prompt{So viel kostet ein Hotel in Chester}{How much does a hotel in Chester cost?} \\
    & Prompt 4 & \prompt{"... alle Mitarbeiter, die alles tun, um hilfsbereit zu sein und sehr freundlich zu sein; köstliche Margaritas; Kolibris und Granatäpfel im Garten (sowie eine sehr freundliche Katze); Ein echtes Gefühl von Zuhause. " Aktionsangebot}{"... all staff, who go out of their way to be helpful and are extremely welcoming; delicious margaritas; hummingbirds and pomegranates in the garden (as well as a very friendly cat); a real home-from-home feeling. "} \\
    & Prompt 5 & \prompt{Gansevoort Land zum Verkauf}{Gansevoort Land for Sale} \\
    \bottomrule
    \end{tabular}
    
    \subcaption{Example where \kNN outperforms random selection.}
    \label{tab:example_kNNWins}
    \end{subtable}
    
    \vspace{1em}
   
    \begin{subtable}{\linewidth}
    \centering \small
    \begin{tabular}{llp{13.3cm}}
    
    \toprule
    & Source    & Frei von Drogen veröffentlichte Green mit der Peter Green Splinter Group einige Alben, bis sich die Band 2004 auflöste. \\
    & Reference &  Free of drugs, Green and the Peter Green Splinter Group released various albums before the band split up in 2004. \\
    \midrule
    
    \multirow{15}{*}{\rotatebox{90}{\kNN \roberta}}
    & Hyp & The band released their debut album, The Last of the Great Pretenders, in 2003. \\
    & Prompt 1 & \prompt{Ab 1990 war er Sänger der Gruppe Talisman, mit der er sieben Studioalben veröffentlichte, bis sich die Band 2007 auflöste.}{From 1998 until his departure in 2007, he was the lead singer of the group Lonestar, which recorded seven studio albums on BNA Records during his tenure as lead vocalist.} \\
    & Prompt 2 & \prompt{2001 veröffentlichte die Band unter dem Namen Glass die rockige Single Out Of Nowhere, verpasste die Charts und löste sich im Anschluss auf.}{Around this time he wrote and presented the ITV Network productions The Rock that Doesn't Roll and The Rock That Rolled Away.} \\
    & Prompt 3 & \prompt{Mit ihrer Band Ex Cops veröffentlichte sie zwei Alben, bevor sich die Band 2015 auflöste.}{Their new band released two EPs before signing to Apparition Records in 2011.} \\
    & Prompt 4 & \prompt{In seiner Jugend gründete David Haering die Punk-Band Side Effect, mit der er drei Alben und eine EP veröffentlichte.}{Peter Hajba and Alexander Brandon used OpenMPT to compose the soundtracks for Bejeweled 2, Bejeweled 3 and other PopCap games.} \\
    & Prompt 5 & \prompt{Nach der Veröffentlichung des Live-Albums Beast from the East 1988 trennten sich die Wege der Musiker, als Don Dokken die Band auflöste.}{In 1988, after the Monsters of Rock Tour and a further platinum album, Don Dokken decided to break up the band and they went on their separate ways.} \\
    
    \midrule
    \multirow{13}{*}{\rotatebox{90}{Random}}
    & Hyp & Free from drugs, Green released several albums with the Peter Green Splinter Group, until the band broke up in 2004. \\
    & Prompt 1 & \prompt{250 gr/m2: eine Reihe merino intermedia, vielseitigkeit und schutz garantiert.}{250 gr/m2: Range merino intermediate, versatility and guaranteed protection.} \\
    & Prompt 2 & \prompt{127 Moo.3, Choeng Thale, Thalang, Phuket, Strand Bang Tao, Thailand (Karte anzeigen)}{127 Moo.3, Choeng Thale, Thalang, Phuket, Bang Tao Beach (Phuket), Thailand (Show map)} \\
    & Prompt 3 & \prompt{Ich bin stolz, sagen zu können, dass Ihr Produkt mir die Größe verliehen hat, von der ich jahrelang geträumt habe.}{I am proud to say that your product has given me the size that I sought for all these years.} \\
    & Prompt 4 & \prompt{In der NBA Saison 2003–2004 schien es endlich besser für Garnett und die Timberwolves zu laufen.}{The 2002–03 season seemed to look up for the Timberwolves.} \\
    & Prompt 5 & \prompt{APC-Fhrancis Oliver Lopez extern natürliche Links}{APC-Fhrancis Oliver Lopez External Passing Juice} \\
    \bottomrule
    \end{tabular}
    \subcaption{Example where random selection outperforms \kNN.}
    \label{tab:example_randomWins}
    \end{subtable}
    
    \caption{Example prompts for random and \kNN \roberta selection for the \langdirtight{German}{English} translation direction.}
    \label{tab:prompt_examples}
    
    \vspace{-1ex}  
\end{table*}

\section{Example Translations}
\label{sec:example-translations}

Tables~\ref{tab:deen_examples} and \ref{tab:enzh_examples} show example translations for \langdirtight{German}{English} and \langdirtight{English}{Chinese} as described in section~\ref{sec:example_translations}.

\begin{table*}
    \begin{subtable}{\linewidth}
    \centering \small
    \begin{tabular}{lp{13.7cm}}
    \toprule
 Source & Wegen eines ersten offiziellen Corona-Verdachtsfalls im Land hat Nordkorea aus Furcht vor einer Katastrophe eine Großstadt komplett abgeriegelt. \\
 Reference &  Due to a first official suspected case of coronavirus in the country, North Korea has completely shut down a major city out of fear of a disaster. \\
    \midrule \multirow{2}{*}{SOTA} & Due to a first official suspected case of coronavirus in the country, North Korea has completely sealed off a major city for fear of a catastrophe. \\ 
    \multirow{2}{*}{PaLM} & North Korea has completely sealed off a city of \major{300,000 people} because of a first suspected case of coronavirus in the country, fearing a catastrophe. \\
    \bottomrule
    \end{tabular}
    
    \subcaption{PaLM generated two errors: \begin{enumerate*}[(i)] \item Makes the assumption that the major city has 300,000 citizens, \item Forgot to translate the word official. SOTA generates a perfect translation. Note: this example also exemplifies PaLM's less literal translation style.\end{enumerate*}}
    \label{tab:deen1}
    \end{subtable}
    
    \vspace{1em}
   
    \begin{subtable}{\linewidth}
    \centering \small
    \begin{tabular}{lp{13.7cm}}
    
    \toprule
    Source    & Für mindestens drei Menschen endet die Tour mit dem Motorrad tödlich. \\
    Reference & For at least three people, their bike ride ended in death.
 \\
     \midrule 
     SOTA & The motorcycle tour ends fatally for at least three people. \\
     PaLM & At least three people \major{die in motorcycle accidents}. \\
    \bottomrule
    \end{tabular}
    \subcaption{The source mentions a single accident happening on a bike tour. PaLM refers to multiple accidents happening independently.}
    \label{tab:deen2}
    \end{subtable}
    
    \vspace{1em}
   
    \begin{subtable}{\linewidth}
    \centering \small
    \begin{tabular}{lp{13.7cm}}
    
    \toprule
    Source    & Ein Zeuge hörte gegen 3.40 Uhr Geräusche in der Talstraße und lief in Richtung des Imbisses.
 \\
    Reference & One witness heard noises on Talstraße around 3:40 am and ran in the direction of the snack stand.
 \\
     \midrule 
     SOTA & A witness heard noises in the \major{valley road} at around \minor{3.40} a.m. and ran towards the snack bar. \\
     PaLM & A witness heard noises in Talstraße at around 3:40 a.m. and ran towards the snack bar. \\
    \bottomrule
    \end{tabular}
    \subcaption{SOTA generates an overly-literal translation, resulting in copying the street name (Talstrasse) and using the wrong time format.}
    \label{tab:deen3}
    \end{subtable}    

    \caption{Example translations from newstest2021 \langdirtight{German}{English}. \palm translations are generated with the high-end prompt pool. These are typical of error patterns observed in the translation output. We also observed the same pattern when using \wmtdev as the prompt pool. In general, SOTA is more faithful to the source while \palm generates less literal translations that occasionally miss some information from the source.}
    \label{tab:deen_examples}
    
    \vspace{-1ex}  
\end{table*}

\begin{table*}



    \begin{subtable}{\linewidth}
        \centering \small
        \begin{tabular}{lp{13.7cm}}
            \toprule
            Source    & French World Cup winner Dembele, who has struggled for game time at the Camp Nou, was recently linked with a move to PSG in a swap deal with Neymar. \\
            Reference & \zh{在诺坎普球场冲锋陷阵的法国世界杯冠军得主 Dembele 最近通过与 Neymar 交换转投 PSG。}                                                                                             \\
            \midrule
            SOTA      & \zh{法国世界杯冠军登贝莱在诺坎普一直在为比赛时间而挣扎，最近他与内马尔交换转会巴黎圣日尔曼。}                                                                                                    \\
            \palm     & \zh{法国世界杯冠军\minor{德容}，在诺坎普的出场时间一直不多，最近被传与内马尔进行交换\major{加钱}转会到PSG。}                                                                                   \\
            \bottomrule
        \end{tabular}
        \subcaption{\palm produces two errors: \begin{enumerate*}[(i)] \item translates a wrong player's name; \item adds extra information that the player received a raise in the swap deal.\end{enumerate*} SOTA produces a perfect translation, but is much more literal than \palm.}
        \label{tab:enzh1}
    \end{subtable}

    \vspace{1em}

    \begin{subtable}{\linewidth}
        \centering \small
        \begin{tabular}{lp{13.7cm}}

            \toprule

            Source    & \dots in the wake of September 11, ASIO was given power to compulsorily question people for up to seven days in relation to terrorism offences. \\
            Reference & \dots \zh{在 911 事件之后，澳安全情报局有权对牵涉恐怖主义行为的人员进行为期最高 7 天的强制性询问。}                                                                                     \\
            \midrule
            SOTA      & \dots \zh{在\major{9月11日}之后，安全情报组织被授权对与恐怖主义罪行有关的人进行长达7天的强制性讯问。}                                                                                  \\
            \palm     & \dots \zh{澳大利亚安全情报局在9·11恐怖袭击之后获得了强制询问人员的权力，可以在7天内就恐怖主义罪行进行询问。}                                                                                  \\
            \bottomrule
        \end{tabular}
        \subcaption{The source phrase ``\textit{September 11}'' is translated literally by SOTA into a date, whereas \palm produces a more appropriate translation by describing it as a terrorist attack.}
        \label{tab:enzh2}
    \end{subtable}

    \caption{Example translations from newstest2021 \langdirtight{English}{Chinese}. \palm translations are generated with the \wmtdev prompt pool. We find SOTA to generate more literal translations than \palm, but \palm suffers from more omissions and mistranslations.}
    \label{tab:enzh_examples}

    \vspace{-1ex}  
\end{table*}

\section{Overlap Analysis}
\label{sec:overlap_analysis}

\begin{table*}
\centering
\begin{tabular}{lclrrrrrr}
\toprule
\multirow{2}[2]{*}{Data} & \multirow{2}[2]{*}{\%Clean} & \multirow{2}[2]{*}{Method} & \multicolumn{3}{c}{\bleu}  & \multicolumn{3}{c}{\bleurt}  \\ \cmidrule(lr){4-6} \cmidrule(lr){7-9}
& & & Orig & Clean & $\neg$Clean & Orig & Clean & $\neg$Clean \\ \midrule
\multirow{3}{*}{\langdir{de}{en} 2016} & \multirow{3}{*}{80.3} & \gtransshort   & 47.6 & 45.5  & 55.3        & 78.4 & 77.7  & 81.3        \\
                                        &                       & \wmtfull Random & 46.1 & 43.5  & 54.9       & 77.3 & 76.3  & 81.5        \\
                                        &                       & Diff           & 1.5  & 2.0   &  0.4        & 1.1  & 1.4   & -0.2        \\ \midrule
\multirow{3}{*}{\langdir{fr}{en} 2014} & \multirow{3}{*}{69.2} & \gtransshort   & 43.1 & 42.1  & 44.8        & 77.7 & 76.8  & 79.6        \\
                                        &                       & \wmtdev Random & 43.0 & 41.3  & 45.4        & 77.7 & 76.9  & 79.5        \\
                                        &                       & Diff           & 0.1  & 0.8   & -0.6        & 0.0  & -0.1  &  0.1        \\
\bottomrule
\end{tabular}

\caption{Comparison between Google Translate and 5-shot \palm using three test sets: Orig. (original), Clean (overlapping examples removed) and $\neg$Clean (including only overlapping examples). We use Random instead of \wmtdev Random for \langdirtight{de}{en} to avoid using the WMT 2021 development sets to prompt for the WMT 2016 test (``sampling from the future'').\label{tab:overlap_compare}}
\end{table*}

\citet{chowdhery2022palm} show \bleu differences between clean and original test sets, 
and provide some evidence that differences are not due to memorization, but it still isn't clear how much overlap actually inflates a model's score.
We directly quantify the effect of train-test overlap on decision making by comparing 5-shot \palm to Google Translate (GT)\footnote{We chose Google Translate for comparison because it is non-trivial to build a SOTA baseline for older WMT scenarios. Through personal communication, we understand that Google Translate has no overlap with WMT test sets.} on
our two sets with substantial overlap, testing under original, clean and $\neg$clean (including only overlapping examples) scenarios.
\bleu and \bleurt scores for the two systems and three test sets are shown in Table~\ref{tab:overlap_compare}.

We can see that directly comparing original and clean results for a single system conflates differences from overlap with those from the increased difficulty of the clean subset.
For example, for \langdirtight{de}{en} \bleu, comparing \palm's original and clean scores gives an overlap gap of 2.6-\bleu, in line with the gaps reported by \citet{chowdhery2022palm}.
However, the non-overlapping GT system also has lower scores on the clean set, indicating that it may simply be more difficult.%
\footnote{The difference in difficulty between Clean and $\neg$Clean for systems without overlap is not easily explained. A common difficulty indicator is sentence length, but average lengths, as measured by number of \sacrebleu tokens per sentence, are similar between Clean and $\neg$Clean for both \langdirtight{de}{en} (23.8 versus 23.0) and \langdirtight{fr}{en} (21.1 versus 22.7).}
It's more useful to see that the original test indicated a 1.5-\bleu difference between the two systems, 
while the clean test indicates a 2.0-\bleu difference, meaning \palm benefited from overlap by 0.5 \bleu in this comparison.
The fully overlapping $\neg$clean further distorts the difference between the two systems: the true (clean) delta of 2.0 \bleu shrinks to only 0.4.
Trends for \langdirtight{fr}{en} are similar: though \palm and GT are very close according to the original test set, the clean set reveals a delta of 0.8 \bleu.
Interestingly, \bleurt may be less sensitive to overlap, with the original-versus-clean deltas hovering around 0 for \langdirtight{fr}{en} regardless of the test subset,
and \langdirtight{de}{en} showing that \palm benefits from an overlap bonus of only 0.3 \bleurt.

In summary, overlap between the target side of the test data and the LLM training data can have an impact on both \bleu and \bleurt scores, altering the delta between two systems where one benefits from overlap and another does not by up to 0.7 \bleu or 0.3 \bleurt for a 20-30\%-overlap.
However, we should emphasize that the differences due to overlap are small overall, and certainly much smaller than expected if one looked only at the difference between original and clean scores.

\section{Fixed versus random prompts}

\begin{table}
    \newcommand{\best}[1]{\textbf{#1}}
    \centering
    \begin{tabular}{llrrr}
    \toprule
     \multirow{2}[2]{*}{LP} & \multirow{2}[2]{*}{Selection} & \multicolumn{3}{c}{\makebox[6ex]{\bleurt}} \\
    \cmidrule(lr){3-5} & & min & avg & max \\
    \midrule

    \multirow{2}{*}{\langdir{en}{de}}
        & fixed         & & 74.7 & \\
        & random     & 74.5 & 74.7 & \best{75.0} \\
    \cmidrule{2-5}
    \multirow{2}{*}{\langdir{de}{en}}
        & fixed         & & \best{76.3} & \\
        & random     & 75.6 & 75.8 & 75.9 \\
    \cmidrule{2-5}
    \multirow{2}{*}{\langdir{en}{zh}}
        & fixed         & & \best{64.7} & \\
        & random     & 63.7 & 63.9 & 64.0 \\
    \cmidrule{2-5}
    \multirow{2}{*}{\langdir{zh}{en}}
        & fixed         & & 67.0 & \\
        & random     & 67.3 & 67.5 & \best{67.7} \\
    \cmidrule{2-5}
    \multirow{2}{*}{\langdir{en}{fr}}
        & fixed         & & \best{75.5} & \\
        & random     & 75.2 & 75.2 & 75.3 \\
    \cmidrule{2-5}
    \multirow{2}{*}{\langdir{fr}{en}}
        & fixed         & & \best{77.9} & \\
        & random     & 77.4 & 77.6 & 77.6 \\
        
    \bottomrule
        
    \end{tabular}
    \caption{
    Fixed (maximum-likelihood) prompts vs random prompts. All prompts are drawn from the \highend pool, and performance is measured on the standard test sets (WMT21 for German and Chinese, WMT14 for French). The scores for random selection are the minimum, average, and maximum over 5 random draws.
    }
    \label{tab:fixed_vs_random}
\end{table}

The results from section~\ref{sec:full-results} indicate that random selection from small, high-quality prompt pools can work better than trying to customize prompts for specific inputs. In this section we investigate the effect of using a \emph{single} high-quality prompt for all inputs, chosen using a maximum-likelihood criterion. For convenience, we carried out experiments on the \highend pool with 1-shot paragraph prompts. For each prompt in the pool, we computed the probability of a set of held-out \highend paragraphs when \palm was conditioned on that prompt. We select the prompt that resulted in the highest probability for each language pair.

Table~\ref{tab:fixed_vs_random} compares this method to random selection from the \highend pool. For all language pairs except \langdirtight{Chinese}{English}, the fixed prompt does as well or better than the average performance over 5 random runs where a different prompt is selected for each input during each run. 
In \langdirtight{Chinese}{English}, the prompt that ranked 5th according to the probability criterion also outperformed the random average, suggesting problems with our held-out set for that language pair.

We conclude that using a single high-quality prompt can be a safer strategy than choosing a fresh randomly-selected prompt for each input. Model probability appears to be a reasonable criterion for judging quality, but we look forward to refining this heuristic in future work.

\end{document}